\documentclass[10pt,twocolumn,letterpaper]{article}

\usepackage{cvpr}
\usepackage{times}
\usepackage{epsfig}
\usepackage{graphicx}
\usepackage{amsmath}
\usepackage{amssymb}

\usepackage{subcaption}
\usepackage{multicol}
\usepackage{multirow}

\begin{document}

%%%%%%%%% TITLE
\title{Explaining, Analyzing, and Probing Representations of Self-Supervised Learning Models for Sensor-based Human Activity Recognition}

\author{Bulat Khaertdinov and Stylianos Asteriadis\\
Maastricht University, Netherlands\\
{\tt\small \{b.khaertdinov,stelios.asteriadis\}@maastrichtuniversity.nl}
% \and
% Stylianos Asteriadis\\
% % First line of institution2 address\\
}

\maketitle
\thispagestyle{empty}

%%%%%%%%% ABSTRACT
\begin{abstract}
   In recent years, self-supervised learning (SSL) frameworks have been extensively applied to sensor-based Human Activity Recognition (HAR) in order to learn deep representations without data annotations. While SSL frameworks reach performance almost comparable to supervised models, studies on interpreting representations learnt by SSL models are limited. Nevertheless, modern explainability methods could help to unravel the differences between SSL and supervised representations: how they are being learnt, what properties of input data they preserve, and when SSL can be chosen over supervised training. In this paper, we aim to analyze deep representations of two recent SSL frameworks, namely SimCLR and VICReg. Specifically, the emphasis is made on (i) comparing the robustness of supervised and SSL models to corruptions in input data; (ii) explaining predictions of deep learning models using saliency maps and highlighting what input channels are mostly used for predicting various activities; (iii) exploring properties encoded in SSL and supervised representations using probing. Extensive experiments on two single-device datasets (MobiAct and UCI-HAR) have shown that self-supervised learning representations are significantly more robust to noise in unseen data compared to supervised models. In contrast, features learnt by the supervised approaches are more homogeneous across subjects and better encode the nature of activities.
\end{abstract}

%%%%%%%%% BODY TEXT

\section{Introduction}
\label{sec:intro}
Sensor-based Human Activity Recognition (HAR) aims to recognize daily activities through signals from wearable devices. Recent sensor-based HAR works rely on Deep Neural Networks trained with large amounts of labeled data. Since data annotation is an expensive process, researchers explore opportunities to train models in an unsupervised, or self-supervised learning, fashion. The SSL training routine normally contains two stages. First, a deep feature encoder is pre-trained using unlabeled data with the help of handcrafted auxiliary task annotations. Secondly, a simple output model is fine-tuned on top of the features learnt during pre-training using available annotated data. While different SSL methods have been adapted to sensor-based HAR \cite{Saeed_2019_multitask, Haresamudram_2020_maskedrec}, contrastive learning has shown the most impressive performance \cite{khaertdinov_dynamic, Jain_2022_Collossl}. In contrastive learning, models are trained to pull together views of input data corresponding to the same instance, also known as positive pairs, while pushing apart negative views coming from different instances. The positive pairs do typically correspond to the different views of the same instance and they can be mined using random augmentations or by extracting different parts of the same signal, e.g. different in time \cite{haresamudram_2020_cpchar, Chen_2020_simclr}. The negative pairs, in turn, are coming from different instances within the batch or from the pre-computed storage of embeddings. The use of negative pairs has been criticized in recent literature \cite{grill_2020_byol, Chen_2021_simsiam} and novel methods alleviating the issues associated with the negative pairs have been proposed \cite{bardes2022vicreg, zbontar_2021_barlow}. 

% In this paper, we adopt one of the novel dimension-contrastive frameworks, namely VICReg \cite{bardes2022vicreg}, that do not require negative pairs to the problem of Human Activity Recognition.

\begin{figure*}[t]
\centering
\begin{minipage}[b]{0.55\linewidth}
  \centering
  \centerline{\includegraphics[width=8.2cm]{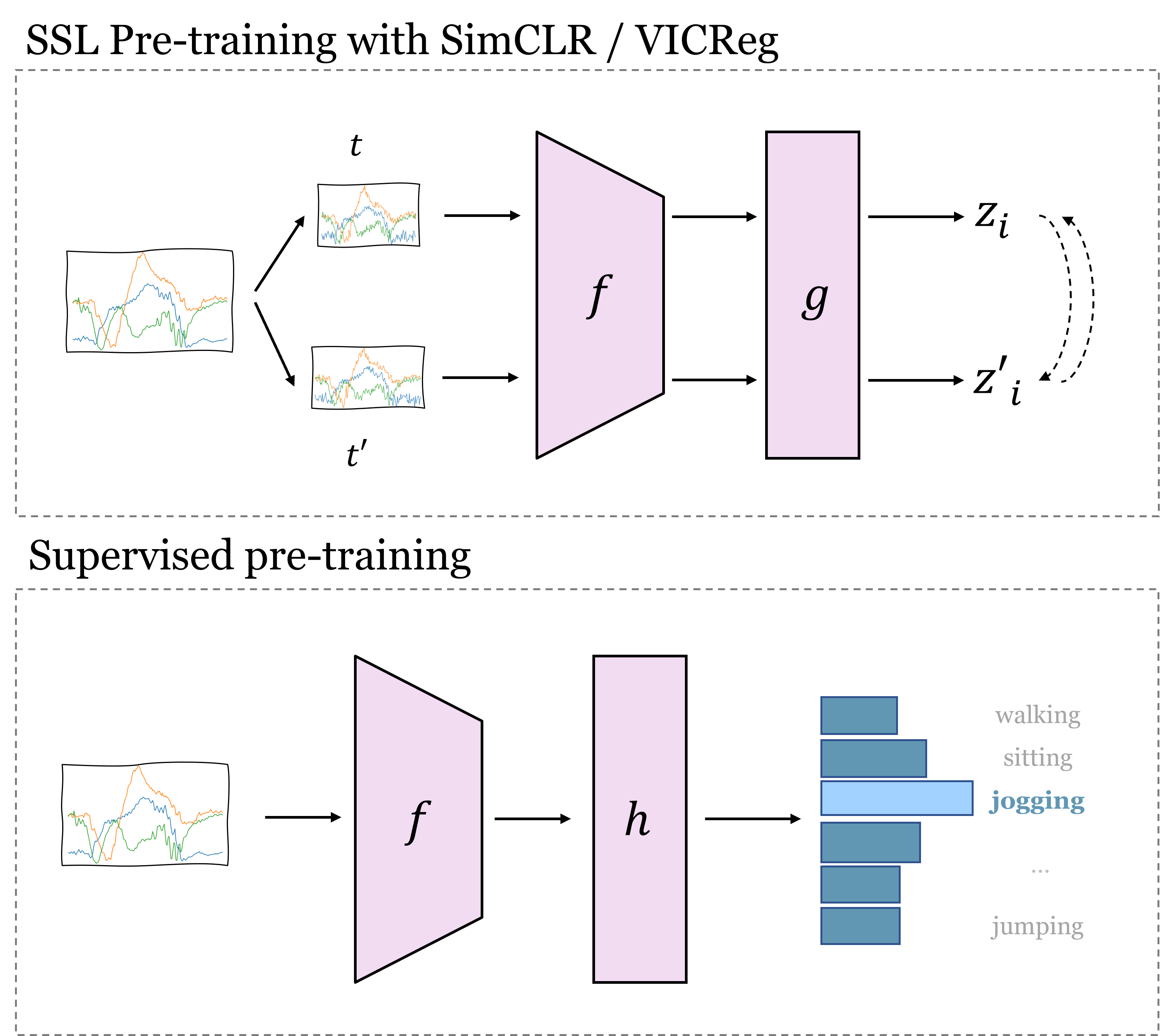}}
%  \vspace{2.0cm}
  \centerline{(a) Pre-training routines}\medskip
  \label{fig:pre-training}
\end{minipage}
\begin{minipage}[b]{0.43\linewidth}
  \centering
  \centerline{\includegraphics[width=6.9cm]{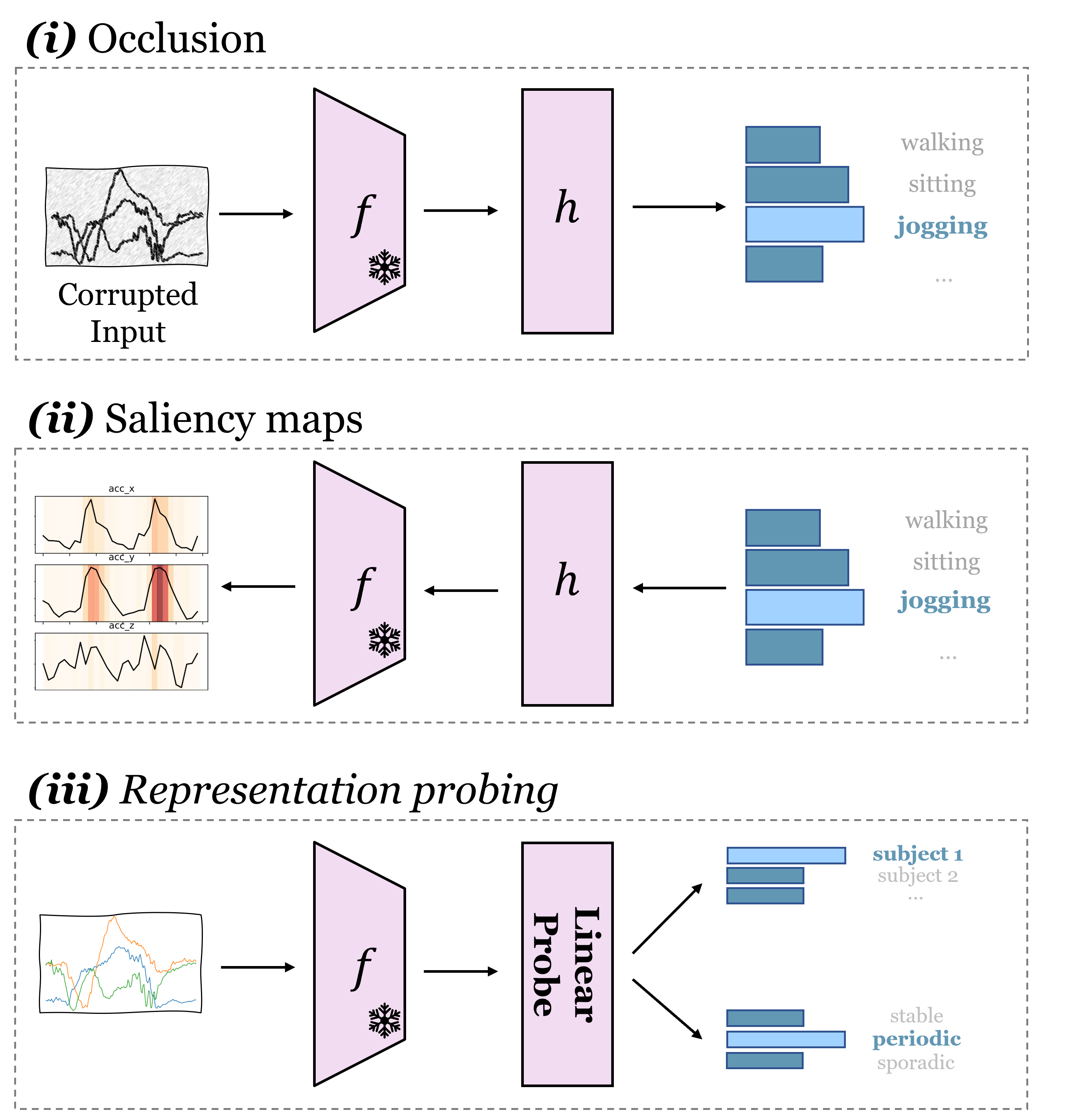}}
%  \vspace{1.5cm}
  \centerline{(b) Feature representation analysis}\medskip
  \label{fig:xai}
\end{minipage}
\caption{The proposed methodology. (a) First, the SSL and supervised models are used to pre-train the encoder $f$. In the case of the supervised model, the classification model $h$ is also pre-trained together with the encoder on annotated data. In contrast, for SSL, the encoder is pre-trained on unannotated data and later the classification model is fine-tuned separately with the frozen encoder. (b) Later, the representations learnt by the encoders are analyzed using three types of explainability methods.}
\label{fig:xai_meth}
\end{figure*}

While different supervised and SSL methods have been successfully adapted to HAR in the past few years, far too little attention has been paid to analyzing representations produced by these algorithms. Nevertheless, this type of analysis can potentially be useful to understand what input channels supervised and SSL frameworks are mostly focused on while making predictions. Besides, it is also important to test the robustness of pre-trained models in realistic cases when input sensory data is corrupted. Finally, exploring the properties of input data that are present in features can be helpful in identifying specific use cases that are more suitable for SSL models. Thus, this paper aims to compare feature representations learnt by SSL and supervised models in different scenarios using various types of explainability methods. Specifically, the main contributions of the paper are listed as follows:
\begin{itemize}
    \item We implement a Transformer-based encoder from \cite{Khaertdinov_2021_csshar} and design one supervised and two SSL training routines (Figure \ref{fig:xai_meth}a). The first SSL framework is an adaptation of the SimCLR contrastive learning framework \cite{Chen_2020_simclr}, or CSSHAR. The second SSL approach is a novel feature decorrelation SSL framework, namely VICReg \cite{bardes2022vicreg}, that does not rely on negative pairs and, as far as the related literature suggests, has not been applied to sensor-based HAR yet.
    \item We analyze representations of the supervised and SSL models using three families of explainability techniques for deep learning topologies shown in Figure \ref{fig:xai_meth}b. First, occlusion experiments have been conducted to compare the robustness of the implemented frameworks to noise in input data. Moreover, Guided Grad-CAM \cite{selvaraju_2017_gradcam}, has been used to obtain local and global saliency maps for multichannel input signals. Finally, representation probing has been adapted to analyze properties of representations, such as subject heterogeneity and activity type.
\end{itemize}
\section{Related Work}
\subsection{Sensor-based HAR and SSL}
Sensor-based HAR is commonly addressed as a time-series classification problem using input signals from Inertial Measurement Units (IMU). Various architectures can be used to build robust feature representations, including Convolutional \cite{Yang_2015, Ignatov_2018} and Recurrent Neural Networks \cite{Zhao_2018}, as well as approaches that use attention mechanisms \cite{Ma_2019_attnsense}. Recently, plenty of SSL frameworks have been suggested to address sensor-based HAR. While earlier works include transformation-based \cite{Saeed_2019_multitask} and reconstruction approaches \cite{Haresamudram_2020_maskedrec}, the latter works are based on contrastive learning frameworks, such as Contrastive Predictive Coding (CPC) and SimCLR \cite{haresamudram_2020_cpchar, khaertdinov_dynamic, Jain_2022_Collossl}, where models are trained to pull together augmented views of input data. These data correspond to the same instance, also known as positive views. The positive pairs are pulled together while pushing apart negative views coming from different instances. The use of negative pairs, though, is associated with several drawbacks. In particular, frameworks exploiting negative pairs do normally require large batch sizes to get enough informative negatives and suffer from false negative pairs, or class collisions, when samples from the same class for a downstream task are considered negative. Thus, novel ways to address this issue have been proposed \cite{brinzea_2022_cmkm, bardes2022vicreg, zbontar_2021_barlow}. The latest works also suggested using only positive pairs by introducing asymmetric gradient flows, e.g. SimSiam \cite{Chen_2021_simsiam}, or feature decorrelation methods, such as Barlow Twins \cite{zbontar_2021_barlow} or VICReg \cite{bardes2022vicreg}, to avoid representation collapse. In this paper, apart from the SimCLR method that has been previously used in HAR, we employ the dimension-contrastive VICReg framework that does not sample negative pairs \cite{garrido2023duality}.

\subsection{Explaining and Probing Representations}
In this paper, we are focused on examining feature representations learnt by self-supervised learning models and exploring their differences from the features produced by supervised models. Interpreting features generated by deep learning algorithms is a challenging task. Nevertheless, there are various explainable AI (XAI) methods that can be used to interpret and analyze deep representations. 

One family of approaches is based on occlusions that mask out parts of input signals \cite{zeiler2014visualizing}. This method relies on the systematic replacement of different input parts and monitoring the drops in performance. Examples of the recently proposed occlusion strategies include but are not limited to highlighting the importance of different body parts for affect recognition \cite{Ghaleb_2021_skeletonxai} or interpreting deep learning models applied to bio-medical sequential data \cite{resta2021occlusion}. The second promising direction to interpret Deep Learning model predictions is to create local explanations, also known as saliency maps, for the raw data inputs. Saliency-based methods can be based on the gradients of the outputs, including Saliency \cite{simonyan_2014_twostream}, Guided Backpropagation \cite{Springenberg_2014}, and Integrated Gradients \cite{sundararajan2017axiomatic}. Another approach to create saliency attributions is to use Class Activation Maps (CAM) and their variations, such as Grad-CAM and Guided Grad-CAM \cite{selvaraju_2017_gradcam}. Methods based on CAMs require the encoders to contain CNNs extracting features from raw data. Finally, the related literature also suggests various adaptations of the modality-agnostic explainability methods, such as Lime \cite{ribeiro2016should} and SHAP \cite{lundberg2017unified}. However, these methods are normally less efficient in terms of computation as they require applying additional transformations to the inputs. Finally, representation probing is a method to interpret what properties of input signals are encoded in learnt representations. For instance, in NLP, an additional classification model can be built on top of learnt features to predict linguistic properties \cite{conneau_2018_cram}. A similar approach has also been recently adopted for Computer Vision applications to evaluate if feature encoders build semantic knowledge and representations reflect such properties as semantic complexity and consistency \cite{basaj_2021_visualprobing}.

So far, very little attention has been paid to explaining and analyzing deep learning methods for sensor-based Human Activity Recognition. To our knowledge, the related literature suggests two types of works on explaining HAR models trained on motion sensor data. First, a few research studies apply local explainability methods to interpret what channels in motion sensor data are used by deep learning methods \cite{uddin_2021human, Jain_2022_Collossl}. The other works exploiting XAI techniques attempt to create human-understandable interpretations of low-level activities and higher-level behaviors. For example, Arrota et al. \cite{Arrota_2022_dexar} adapted various XAI methods to create activity explanations using wearable and environmental binary sensors in smart-home environments. In recent work, Jeyakumar et al. \cite{Jeyakumar_2023} proposed an end-to-end deep activity recognition model that, apart from detecting simple low-level, is capable of recognizing long-term complex patterns and provides the human-understandable reasoning for them. 

The main objective of this paper is to examine similarities and differences of representations learnt by SSL and supervised models by using various families of explainability methods. Such analysis is conducted to have a better understanding of the differences between SSL and supervised representations, strengths and weaknesses of SSL frameworks, and appropriate use cases. 

% In this paper, we employ explainability methods to conduct different and more low-level
% we employ explainability and representation analysis methods in order to look deeper into feature representations learnt by supervised and SSL methods. 
% Namely, we emphasize the importance of certain devices in input data, evaluate the robustness of the models to noise, and evaluate the properties of data encoded by the models.  

%Uddin et al. \cite{uddin_2021human} used the black box LIME method for assessing the importance of input channels for the supervised LSTM network.
\section{Methodology}
\subsection{Problem Definition and Feature Encoder}
Human Activity Recognition using inertial sensors is normally addressed as multivariate time-series classification. Namely, each input signal from $S$ channels of length $T$ can be written as $\boldsymbol{X} = [\boldsymbol{x}_1, \boldsymbol{x}_2, \ldots, \boldsymbol{x}_T] \in \mathbb{R}^{T \times S}$. An input of the sequence at timestep $t$ can be defined as $\boldsymbol{x}_t = [x_t^1, x_t^2, \ldots, x_t^S] \in \mathbb{R}^{S}$. Each input sequence $\boldsymbol{X}$ is associated with label $y \in Y$, where $Y$ represents a set of activities. 
In this paper, we encode input signals using convolutional layers followed by transformer self-attention blocks as has been previously done in \cite{Khaertdinov_2021_csshar}. In the SSL fine-tuning stage, the obtained features are flattened and passed to a linear layer with softmax activation for activity classification.

Two stages of training are typically implemented in SSL frameworks. First, during the pre-text task, encoder $f: \mathbb{R}^{T \times S} \xrightarrow{} \mathbb{R}^{D}$ is trained to encode input data as a vector of size $D$ on auxiliary task without using downstream task labels. Next, in fine-tuning, the learnt features are passed to a classification model $h: \mathbb{R}^{D} \xrightarrow{} \mathbb{R}^{Y}$ producing values corresponding to softmax probabilities for each activity in a dataset. During SSL pre-training, before computing the loss, features are passed through an additional projection head, typically a shallow MLP, that can be defined as $g: \mathbb{R}^{D} \xrightarrow{} \mathbb{R}^{D'}$ where $D'$ is the dimensionality of the projected representations. In this paper, we use two recent SSL frameworks, namely SimCLR \cite{Chen_2020_simclr} and VICReg \cite{bardes2022vicreg}. 

\subsection{SimCLR} 
In our adaptation of SimCLR, a set of random augmentations is used to create two views of the input instances. Formally, two sets of $Z = [z_1, z_2, \dots, z_N] \in \mathbb{R}^{N \cdot D'}$ and $Z' = [z'_1, z'_2, \dots, z'_n] \in \mathbb{R}^{N \cdot D'}$ are generated by passing augmented views of input instances in a batch of size $N$ to encoder $f(\cdot)$ and projection MLP $g(\cdot)$. Hence, $z_i = g(f(t(\boldsymbol{X}_i))) \in \mathbb{R}^{D'}$ and $z'_i = g(f(t'(\boldsymbol{X}_i))) \in \mathbb{R}^{D'}$ are the representations of the same multivariate time-series $\boldsymbol{X}_i$ under two different augmentations $t$ and $t'$, respectively. Later, two views from the same instance forming a positive pair are contrasted against all the negative pairs in a batch using the Negative Temperature-scaled Cross-Entropy loss function (NT-Xent) \cite{Chen_2020_simclr}:

\begin{equation}
    l(z_i, z'_i) = -log\frac{exp(\frac{s_(z_i, z'_i)}{\tau})}{\sum_{z'_k \in Z'} exp(\frac{s_(z_i, z'_k)}{\tau})},
    \label{eq:NT-Xent}
\end{equation}
where $s(z_i, z'_i)$ is a cosine similarity between projected features $z_i$ and $z'_i$, and $\tau$ is a temperature hyperparameter.

The final cost function for a batch is defined as follows:
\begin{equation}
    L_{c} = \sum_i^N (l(z_i, z'_i) + l(z'_i, z_i)).
    \label{eq:NT-Xent-cost}
\end{equation}

The downside of the SimCLR framework lies in the use of negative pairs. They normally require large batches and form so-called false negative pairs when inputs belonging to the same class of a downstream task are treated as a negative pair. 

\begin{table*}[!t]
\centering
\scalebox{0.75}{
\begin{tabular}{|l|l|l|}
\hline
           & MobiAct                                               & UCI-HAR                                       \\ \hline
Periodic   & Walking, Jogging, Jumping, Stairs Up, Stairs Down     & Walking, Walking Downstairs, Walking Upstairs \\ \hline
Stable     & Standing, Sitting                                     & Sitting, Standing, Laying                     \\ \hline
Transition & Stand to Sit, Sit to Stand, Car-step in, Car-step out & -                                             \\ \hline
\end{tabular}}
\caption{Categorization of activities present in the UCI-HAR and MobiAct datasets. We use activity names provided by the developers of the datasets.}
\label{tab:activity-types}
\end{table*}

\subsection{VICReg} 
The VICReg framework \cite{bardes2022vicreg} does not use negative pairs and, in turn, introduces regularization terms applied to two sets of augmented instances $Z$ and $Z'$ in a batch. Three loss components are used during model pre-training. First, the invariance term is calculated as the mean squared error between positive pairs in $Z$ and $Z'$ as follows:
\begin{equation}
    d(Z, Z') = \frac{1}{N} \sum_i^N ||z_i - z'_i||_2^2.
\end{equation}

The second regularization term assures that variance is maintained at level $\gamma$ over all dimensions in both $Z$ and $Z'$ using hinge loss:
\begin{equation}
    v(Z) = \frac{1}{D'} \sum_i^{D'} \max(0, \gamma - S(z^j)),
\end{equation}
where $S(z^j) = \sqrt{Var(z_j) + \epsilon}$. 

Finally, the last component of the loss function, inspired by \cite{zbontar_2021_barlow}, decorrelates different dimensions of representations in order to prevent them from encoding the same information. Specifically, the off-diagonal elements of the covariance matrices $C(Z)$ and $C(Z')$ are minimized. In this case, the loss for $C(Z)$ can be formulated as follows:
\begin{equation}
    c(Z) = \frac{1}{D'} \sum_{i \neq j} [C(Z)]_{i,j}^2
\end{equation}

As a result, the overall cost function for $Z$ and $Z'$ can be defined as follows \cite{bardes2022vicreg}:
\begin{equation}
    L_{v}(Z, Z') = \lambda d(Z, Z') + \mu [v(Z) + v(Z')] + \nu[(c(Z) + c(Z')],
    \label{eq:vicreg}
\end{equation}
where $\lambda$, $\mu$ and $\nu$ are hyperparameters weighing the importance of each loss component.

\subsection{Analysing Feature Representations}
\label{sec:meth_xai}

In this paper, we adapt three different explainability approaches illustrated in Figure \ref{fig:xai_meth}b to the sensor-based HAR settings and employ them to compare the SSL and supervised feature representations. In particular, we aim to compare the robustness of the implemented methods in realistic scenarios with corrupted data using occlusions (i). Furthermore, we employ Guided Grad-CAM (ii) to obtain local and global attributions that indicate what input channels the implemented methods use the most to predict each activity. Finally, we exploit representation probing (iii) to analyze if the extracted features contain two properties of input data, namely subject information and the nature of activities.

\noindent{\textbf{(i) Occlusion.}}
We occlude data from certain devices using Gaussian noise, in order to explore the importance of these devices in learning meaningful representations as well as to test the robustness of models to noise in unseen test data. Specifically, the signals from occluded channel $\boldsymbol{x^i}$ are sampled from $\mathcal{N}(\mu,\sigma^{2})$, where $\mu=0$ and $\sigma=1$. We employ two occlusion scenarios to test the robustness of the models. First, we occlude both training and test data from either all accelerometer or all gyroscope channels. In this way, we explore the importance of these devices in learning representations. In the second case, we randomly mask out different numbers and combinations of input channels in test data only. This scenario is closer to the real-world setting when the representations are learned on cleansed data but later applied to unseen data that can be partially corrupted.

\noindent{\textbf{(ii) Guided Grad-CAM.}}
Saliency maps are an XAI method that aims to highlight the most important parts of input data for a decision made by a model. There are various approaches for generating saliency maps. The main requirement for the algorithm to be exploited in our case is that it has to produce saliency maps that highlight the most influential parts of signals on both temporal and spatial (sensor) levels. Therefore, we propose using gradient-based Guided Grad-CAM \cite{selvaraju_2017_gradcam} algorithm to create local class-discriminative attributions that indicate the importance of each input channel at a certain timestamp. We prefer using gradient-based model-specific methods because they do not require creating perturbation routines for input data and are less computationally expensive compared to model-agnostic algorithms, such as LIME or SHAP \cite{molnar_2022_xai}. Guided Grad-CAM combines widely used Guided Backpropagation \cite{Springenberg_2014} and Grad-CAM algorithms using element-wise multiplication for the obtained saliency maps. 

\noindent{\textbf{(iii) Representation Probing.}} 
The main purpose of probing is to analyze if learnt representations capture various properties of input data. Such analysis can help to understand why implemented models can fail in certain scenarios, how well representations describe raw input signals, and what use cases can be successfully employed using a certain model. Typically, an additional shallow model is employed on top of the features to predict labels representing a certain property. For sensor-based HAR, we suggest studying if the representations of supervised and SSL models are subject invariant and contain cues about the type of activities. We train a linear classification model with softmax activation for both probing tasks using frozen features as inputs.

One of the challenges in HAR is subject heterogeneity, as people may perform the same activities differently. On the one hand, this phenomenon could negatively affect the final performance on unseen subjects if an implemented model fails to build robust subject-independent representations. On the other hand, learning personalized features could be beneficial for some applications. A subject heterogeneity probing task evaluates the subject invariance of the representations. In other words, a probing model aims to predict subject labels given feature representations of the pre-trained model. Specifically, after pre-training, feature representations are obtained for the whole dataset including validation and test splits. Later, the representations are randomly divided into 5 folds (4 folds for training and 1 for testing). 

The second probing task, namely activity type, explores if learnt representations contain information about the nature of the activity. According to previous studies \cite{bulling_2014_tutorial, shao_2022_lstm_har}, activities can be categorized into 3 groups, namely periodic (active), stable, and sporadic (transition). Periodic activities, such as walking, jogging, going upstairs or downstairs, etc., do normally contain repetitive patterns in movement and, hence, in recorded signals. Stable activities correspond to postures and static activities. Finally, sporadic activities are rarely observed activities that can occur in transitions between periodic and stable activities. Thus, in the activity type probing task, we categorize activities into these three groups. Later, we hold out one activity from each group for testing and train a probing model to recognize activity types using feature representations of the remaining activities. Hence, the test fold contains one previously unseen activity from each type. We repeat this experiment for all possible combinations of activities in the test set. 

\section{Implementation Details}
% We use two single-device containing data from a single IMU with one accelerometer and one gyroscope (6 channels, overall). Hence, we analyze the importance of accelerometers and gyroscopes as well as their individual channels.
% \noindent\textbf{PAMAP2}. The PAMAP2 dataset \cite{Reiss_2012_Pamap} contains data from three wearable devices on the waist, wrist, and ankle of the leading hand and leg, respectively. We follow a widely used protocol extracting heart rate signals as well as accelerometer, gyroscope and magnetometer channels per device (28 channels overall); signals are downsampled to 33.3 Hz and time windows of 5.12 seconds are extracted with 1-second overlap \cite{Hammerla_2016}. Sequences obtained from subjects 5 and 6 were used as validation and test sets, respectively. 
\subsection{Datasets}
We extract the three-axis accelerometer and three-axis gyroscope device data (i.e., 6 channels overall) from MobiAct \cite{vavoulas_2016_mobiact} and UCI-HAR \cite{Anguita_2013_uci} datasets. The UCI-HAR dataset \cite{Anguita_2013_uci} has been acquired using smartphones equipped with IMUs placed on the waists of subjects. Overall, 30 subjects took part in data collection performing the basic 6 activities of daily living. The MobiAct \cite{vavoulas_2016_mobiact} dataset contains data collected from 61 subjects using smartphones placed in their pockets. Overall, 11 daily activities are present in the dataset. For occlusion experiments, we employ a standard protocol adapted from previous studies \cite{haresamudram_2020_cpchar, Khaertdinov_2021_csshar} in the literature. Namely, signal sequences are segmented in 1-second 50\% overlapping time windows. 20\% of the users were randomly held out for the test set, 20\% of the remaining subjects -- for the validation set, and the remaining ones were used for training. We normalize signals to zero mean and unit variance per channel. The categorization of the activities in the datasets into periodic, stable, and sporadic for the activity type probing task is demonstrated in Table \ref{tab:activity-types}.

\subsection{Encoder}
We use a transformer encoder introduced in \cite{Khaertdinov_2021_csshar}. Specifically, raw signals are passed through 3 layers of one-dimensional CNN (1D CNN). The number of output channels is set to $[64, 128, 256]$ and $[32, 64, 128]$ for MobiAct and UCI-HAR, respectively. The kernel size is 3 for both datasets. The obtained feature maps are passed through positional encoding block and transformer self-attention layers. The number of self-attention layers is 6 with 8 heads and 8 with 8 heads for MobiAct and UCI-HAR, respectively.

\subsection{Self-Supervised Learning}
Both SSL frameworks, SimCLR and VICReg, require augmentations applied to input data. Similarly to \cite{Khaertdinov_2021_csshar}, we use random combinations of jittering, scaling and rotation for the MobiAct dataset and jittering, scaling and permutation for UCI-HAR. For SimCLR, we set $\tau = 0.1$; for VICReg, $\lambda=10$, $\mu=10$, $\nu=5$.

We pre-train the encoder with both SimCLR and VICReg for 200 epochs using LARS applied over Adam optimizer with learning rate $10^{-4}$ \cite{you_2017_lars}. The mini-batch sizes used for MobiAct and UCI-HAR are set to 256 and 128, respectively. The projection head is a 2-layer MLP. For MobiAct, the number of neurons in the hidden and output layer in projection is 512 and 256, for UCI-HAR -- 256 and 256.

\subsection{Supervised training and fine-tuning}
During supervised training and fine-tuning, the output of the transformer encoder is flattened and passed through a linear layer with softmax activation. The models are trained for 100 epochs using Adam optimizer with learning rate $10^{-3}$.

\section{Evaluations}

\begin{table}[t]
\centering
\scalebox{0.75}{
\begin{tabular}{r|ccc|ccc}
              & \multicolumn{3}{c|}{MobiAct}                                         & \multicolumn{3}{c}{UCI-HAR}                                          \\
Mask          & Sup                   & SimCLR                       & VicReg & Sup                   & SimCLR                       & VicReg \\ \hline
None          & {83.25} & {81.2}  & 81.56  & {91.23} & {91.93} & 90.2   \\
Acc & {65.11} & {65.19} & 65.36  & {64.39} & {61.07} & 62.24  \\
Gyro     & {79.65} & {71.81} & 73.64  & {89.71} & {88.25} & 86.73
\end{tabular}}
\caption{Average F1-scores for single IMU occlusion experiments. Sup, Acc and Gyro stand for supervised, accelerometer and gyroscope, respectively.}
\label{tab:train_occlusion_single}
\end{table}

\begin{figure}[t]
\centering
\begin{minipage}[b]{0.48\linewidth}
  \centering
  \centerline{\includegraphics[width=3.8cm]{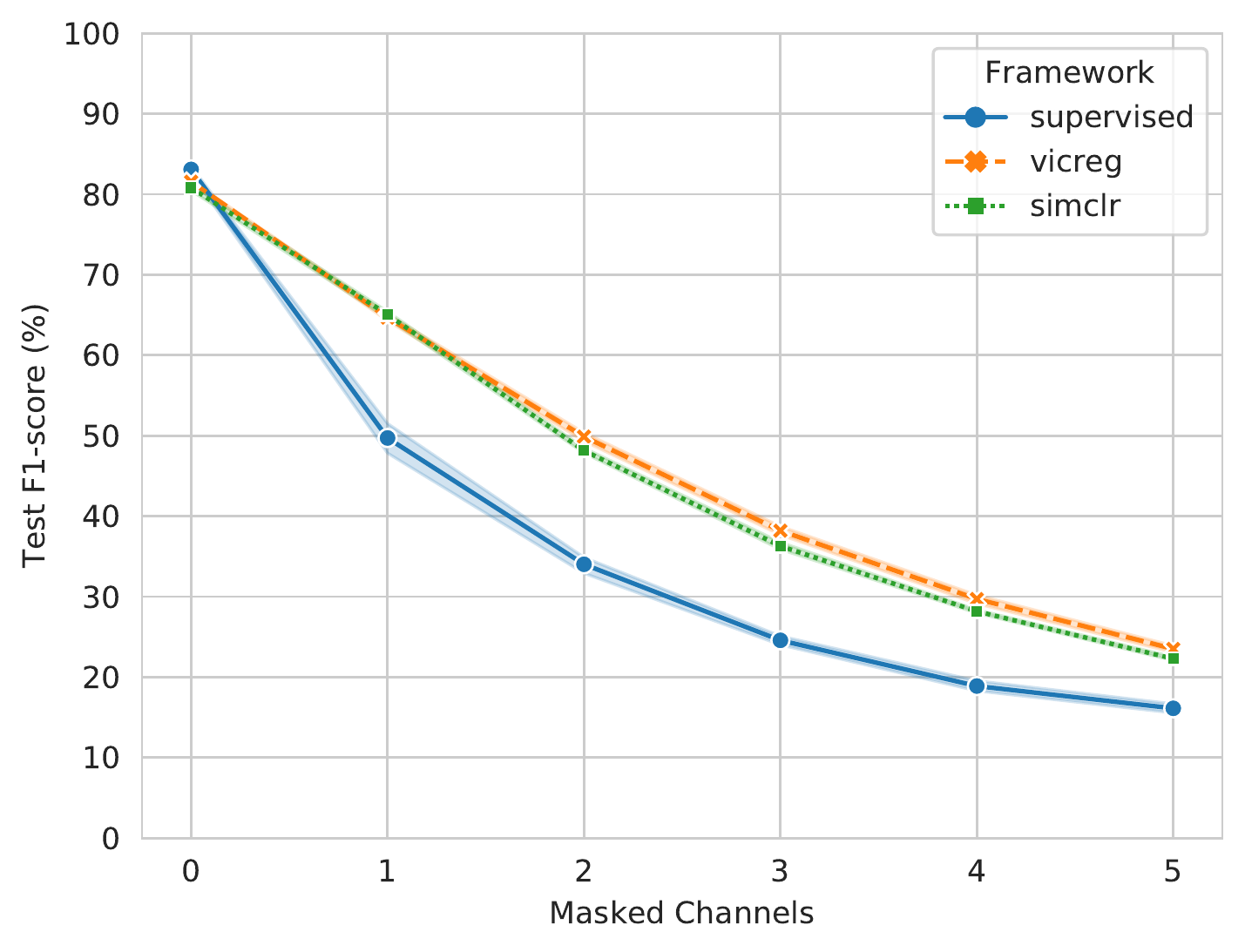}}
%  \vspace{2.0cm}
  \centerline{(a) MobiAct}\medskip
\end{minipage}
\begin{minipage}[b]{0.48\linewidth}
  \centering
  \centerline{\includegraphics[width=3.8cm]{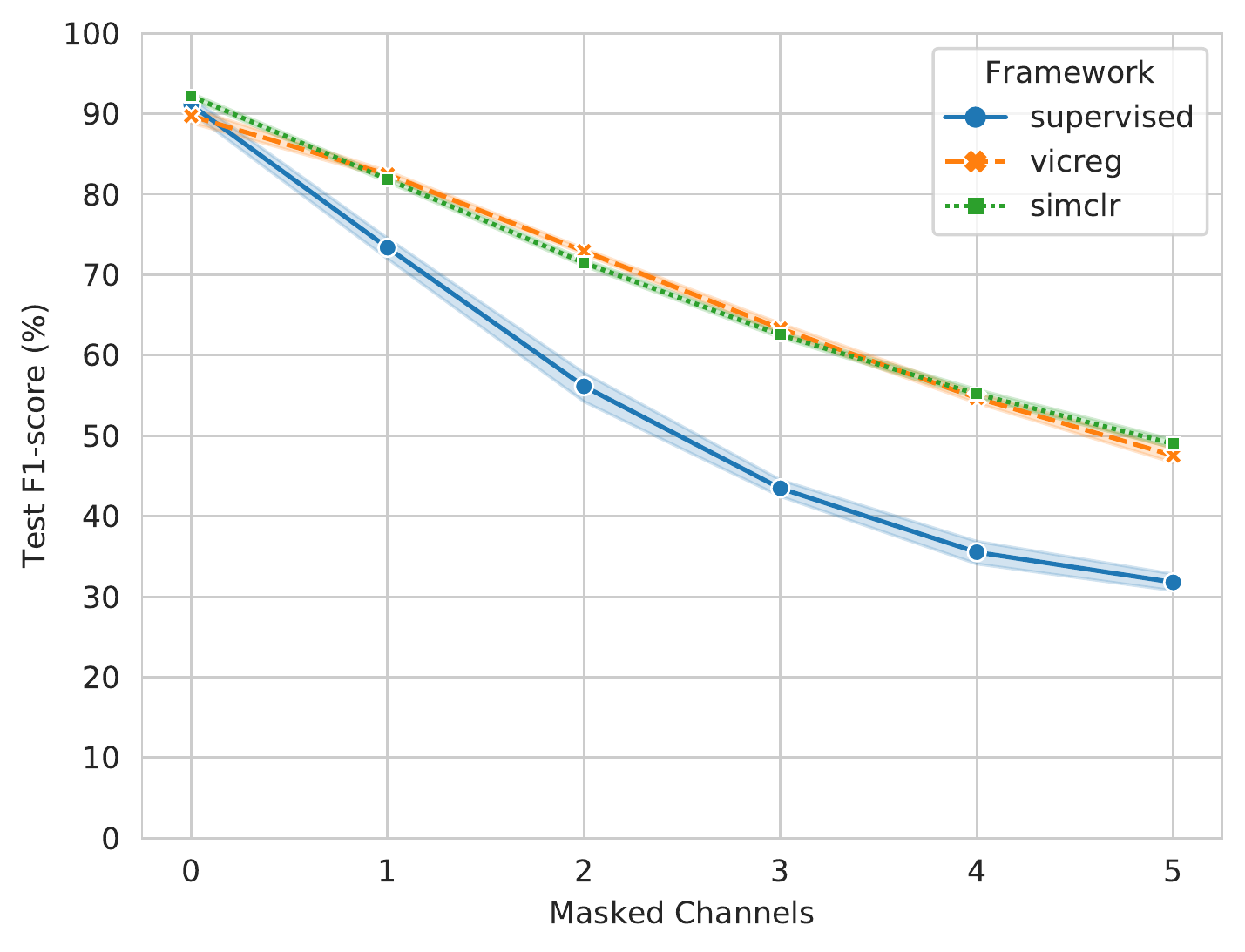}}
%  \vspace{1.5cm}
  \centerline{(b) UCI-HAR}\medskip
\end{minipage}
% \hfill
% \begin{minipage}[b]{.32\linewidth}
%   \centering
%   \centerline{\includegraphics[width=2.9cm]{figures/pamap2_random_test_noise.pdf}}
%   \centerline{(c) PAMAP2}\medskip
% \end{minipage}
%
\caption{The values of test F1-score with 95\% confidence intervals for the random test occlusion experiment.}
\label{fig:test_noise}
\end{figure}

\subsection{Occlusion}
First, we mask input signals out with Gaussian noise during both training and testing. Namely, we separately occlude either accelerometer or gyroscope signal channels to evaluate their importance on learning robust activity representations. In UCI-HAR and MobiAct, both the accelerometer and gyroscope have 3 input channels, meaning that for each experiment we mask out all 3 input channels from one of these devices. We show the obtained macro average F1-scores for this experiment in Table \ref{tab:train_occlusion_single}. As can be seen from the table, all 3 methods experience a large drop in performance when all accelerometer channels are masked out. What is more, SSL models show a larger decline compared to the supervised model when gyroscope data have been masked in the MobiAct dataset.

% In the case of PAMAP2, we mask out each device one by one to evaluate the contribution of body parts to the final performance. Results obtained for these experiments are shown in Tables \ref{tab:train_occlusion_single} and \ref{tab:train_occlusion_pamap}. 

% In contrast, while multiple devices are available, unlike in a fully-supervised model, the SSL model benefits from excluding some body parts. The largest performance boost for both SimCLR and VICReg occurred when signals from IMU placed on the ankle have been masked out (last row in Table \ref{tab:train_occlusion_pamap}).

Additionally, we analyze the robustness of feature representations to random noise in unseen test data only. Specifically, we mask out k $\in [1, 2, 3, 4, 5]$ randomly selected channels for each instance in the test set. We repeat this experiment 10 times with different values of a random seed and show the average F1-score with 95\% confidence intervals in Figure \ref{fig:test_noise}. We also show the results for $k = 0$ when no channels are masked and test data is not altered. The supervised model normally outperforms the SSL models when no noise is applied to unseen data. However, when one or more input channels are masked out, the SSL models show much more robust performance and typically outperform the supervised model by at least 10\%. For UCI-HAR, this gap reaches its peak, namely about 20\%, when 5 out of 6 channels are masked out. 

\begin{table}[!t]
\centering
\scalebox{0.75}{
\begin{tabular}{r|cc|cc|cc}
                           & \multicolumn{2}{c|}{Supervised}             & \multicolumn{2}{c|}{SimCLR}           & \multicolumn{2}{c}{VICReg}            \\
                           & 0$\rightarrow$1 & 1$\rightarrow$2 & 0$\rightarrow$1 & 1$\rightarrow$2 & 0$\rightarrow$1 & 1$\rightarrow$2 \\ \hline
Standing                   & \textbf{53.2}              & \underline{28.6}              & 9.0               & 27.3              & 15.3              & 22.6              \\
Walking                    & \textbf{10.7}              & 13.2              & 5.2               & 10.6              & 6.1               & \underline{13.9}              \\
Jogging                    & \textbf{9.3}               & \underline{12.4}              & 9.0               & 11.2              & 6.9               & 10.4              \\
Jumping                    & \textbf{9.3}               & \underline{6.0}               & 0.5               & 2.3               & 0.6               & 1.2               \\
Stairs up                  & \textbf{26.0}              & \underline{14.7}              & 8.7               & 11.7              & 5.7               & 10.5              \\
Stairs down                & 9.4               & 15.2              & \textbf{13.2}              & \underline{17.4}              & 12.0              & 14.0              \\
Stand to sit & \textbf{42.8}              & \underline{22.6}              & 21.0              & 20.6              & 17.7              & 22.0              \\
Sitting on chair           & \textbf{45.5}              & 14.0              & 36.9              & \underline{19.9}              & 31.7              & 18.0              \\
Sit to stand     & \textbf{49.0}              & 5.9               & 15.9              & \underline{11.3}              & 11.4              & 10.3              \\
Car-step in                & \textbf{12.7}              & \underline{14.0}              & 7.2               & 6.1               & 5.9               & 5.6               \\
Car-step out               & \textbf{21.1}              & \underline{14.1}              & 7.7               & 9.2               & 5.0               & 9.3              
\end{tabular}}
\caption{Average performance drop per activity during the test occlusion experiment for the MobiAct dataset. The largest drop in performance for zero to one and one to two masked channels are highlighted in bold and underlined, respectively.}
\label{tab:mobi-test-occ}
\end{table}

\begin{table}[!t]
\centering
\scalebox{0.75}{
\begin{tabular}{r|cc|cc|cc}
                   & \multicolumn{2}{c|}{Supervised}                                       & \multicolumn{2}{c|}{SimCLR}                                     & \multicolumn{2}{c}{VICReg}                                      \\
                   & 0 $\rightarrow$ 1 & 1 $\rightarrow$ 2 & 0 $\rightarrow$ 1 & 1 $\rightarrow$ 2 & 0 $\rightarrow$ 1 & 1 $\rightarrow$ 2 \\ \hline
Walking            & \textbf{8.7}                            & 6.7                            & 4.6                            & 10.4                           & 5.9                            & \underline{13.6}                           \\
Walking Upstairs   & 10.7                           & 13.0                           & \textbf{11.2}                           & 12.9                           & 10.8                           & \underline{13.7}                           \\
Walking Downstairs & \textbf{10.2}                           & \underline{9.5}                            & 4.7                            & 6.1                            & 3.5                            & 7.3                            \\
Sitting            & \textbf{20.9}                           & \underline{21.4}                           & 5.4                            & 9.7                            & 2.5                            & 2.5                            \\
Standing           & \textbf{37.3}                           & \underline{33.6}                           & 23.1                           & 12.6                           & 10.4                           & 9.3                            \\
Laying             & 15.7                           & 12.8                           & 13.2                           & 10.5                           & \textbf{17.4}                           & \underline{13.5}            
\end{tabular}}
\caption{Average performance drop per activity during the test occlusion experiment for the UCI-HAR dataset. The largest drop in performance for zero to one and one to two masked channels are highlighted in bold and underlined, respectively.}
\label{tab:uci-test-occ}
\end{table}

In order to conduct a deeper analysis of the observed behavior during test time occlusion, we have calculated the absolute difference between the values of recall per activity for 0 and 1, and 1 and 2 randomly masked channels (Tables \ref{tab:mobi-test-occ} and \ref{tab:uci-test-occ}). In other words, the drop in average recall is measured for each activity when the number of masked channels is increased from 0 to 1 (0 $\rightarrow$ 1) and from 1 to 2 (1 $\rightarrow$ 2). The supervised model shows the largest drop in performance for almost all the activities in either 0 $\rightarrow$ 1 or 1 $\rightarrow$ 2 cases. What is more, the decline in the performance of the supervised model is considerably greater compared to the SSL frameworks for stable and transition activities even if only one input channel is masked, especially for standing (53.2\% on MobiAct, 37.3\% on UCI-HAR), sitting (45.5\% on MobiAct, 20.9\% on UCI-HAR), and sit-to-stand and stand-to-sit (49.0\% and 42.8\% on MobiAct). 

\subsection{Local and Global Saliency}
\begin{figure}[t]

\begin{minipage}[b]{.32\linewidth}
  \centering
  \centerline{\includegraphics[width=2.2cm]{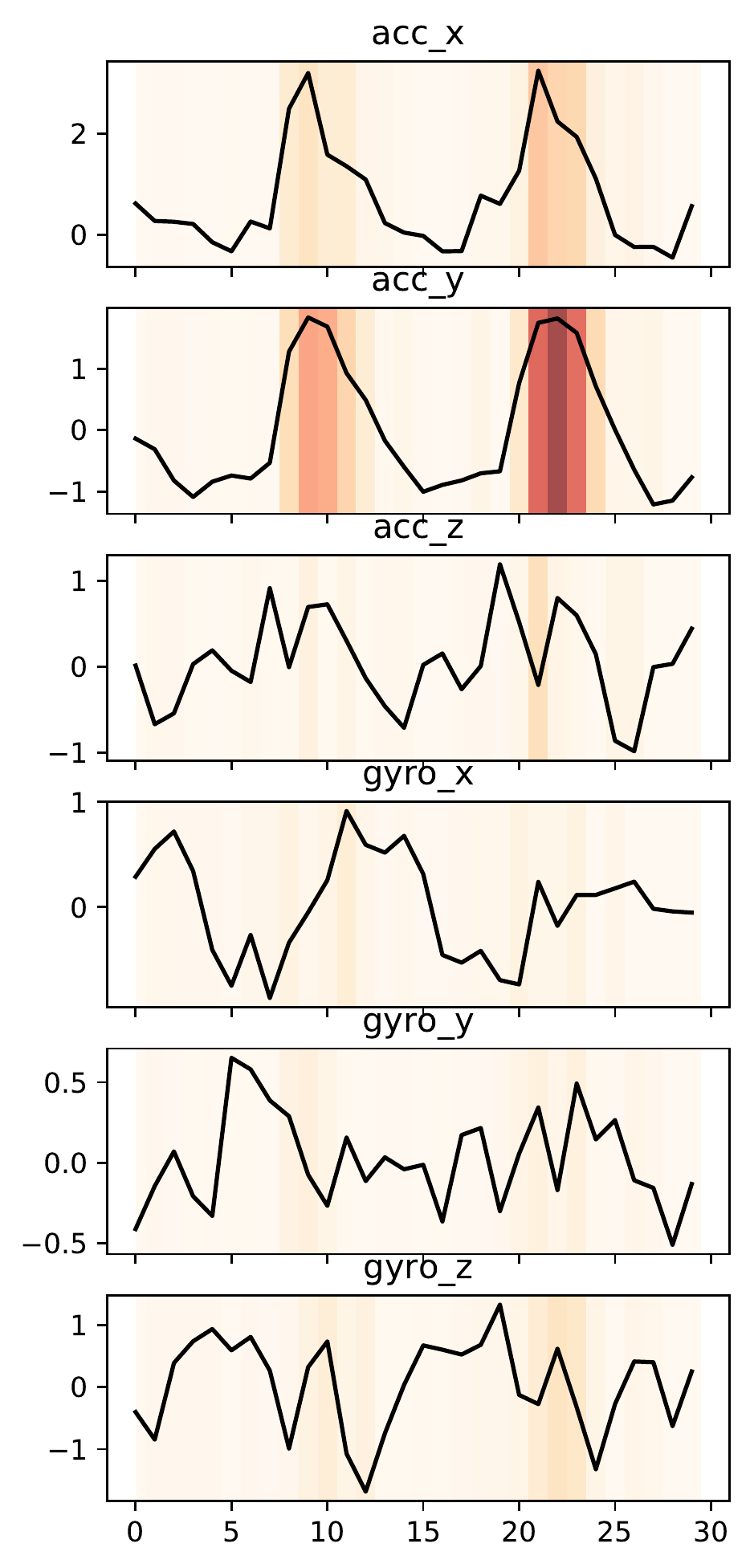}}
%  \vspace{2.0cm}
  \centerline{(a) Sup: Jump}\medskip
\end{minipage}
\hfill
\begin{minipage}[b]{.32\linewidth}
  \centering
  \centerline{\includegraphics[width=2.2cm]{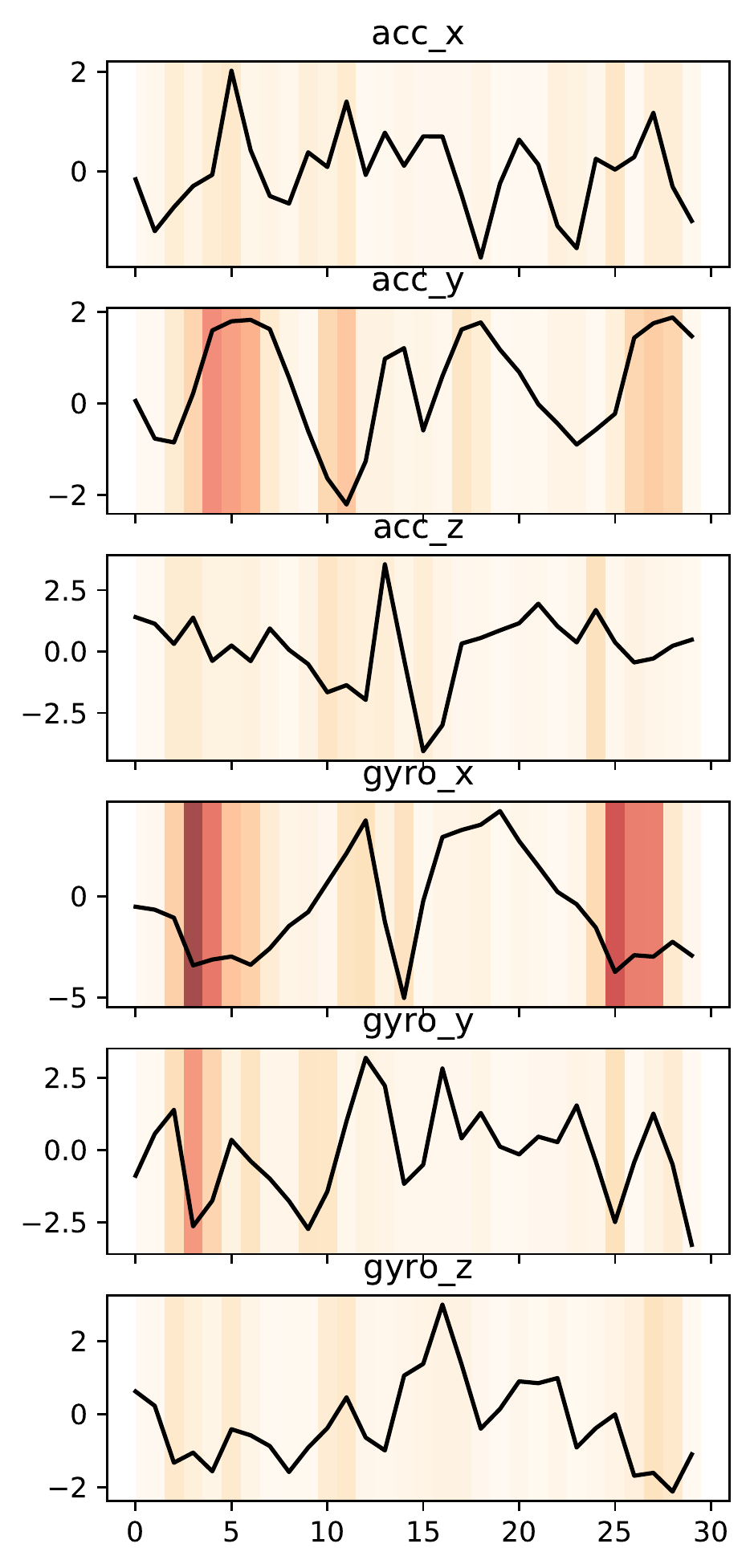}}
%  \vspace{1.5cm}
  \centerline{(b) Sup: Jog}\medskip
\end{minipage}
\hfill
\begin{minipage}[b]{.32\linewidth}
  \centering
  \centerline{\includegraphics[width=2.12cm]{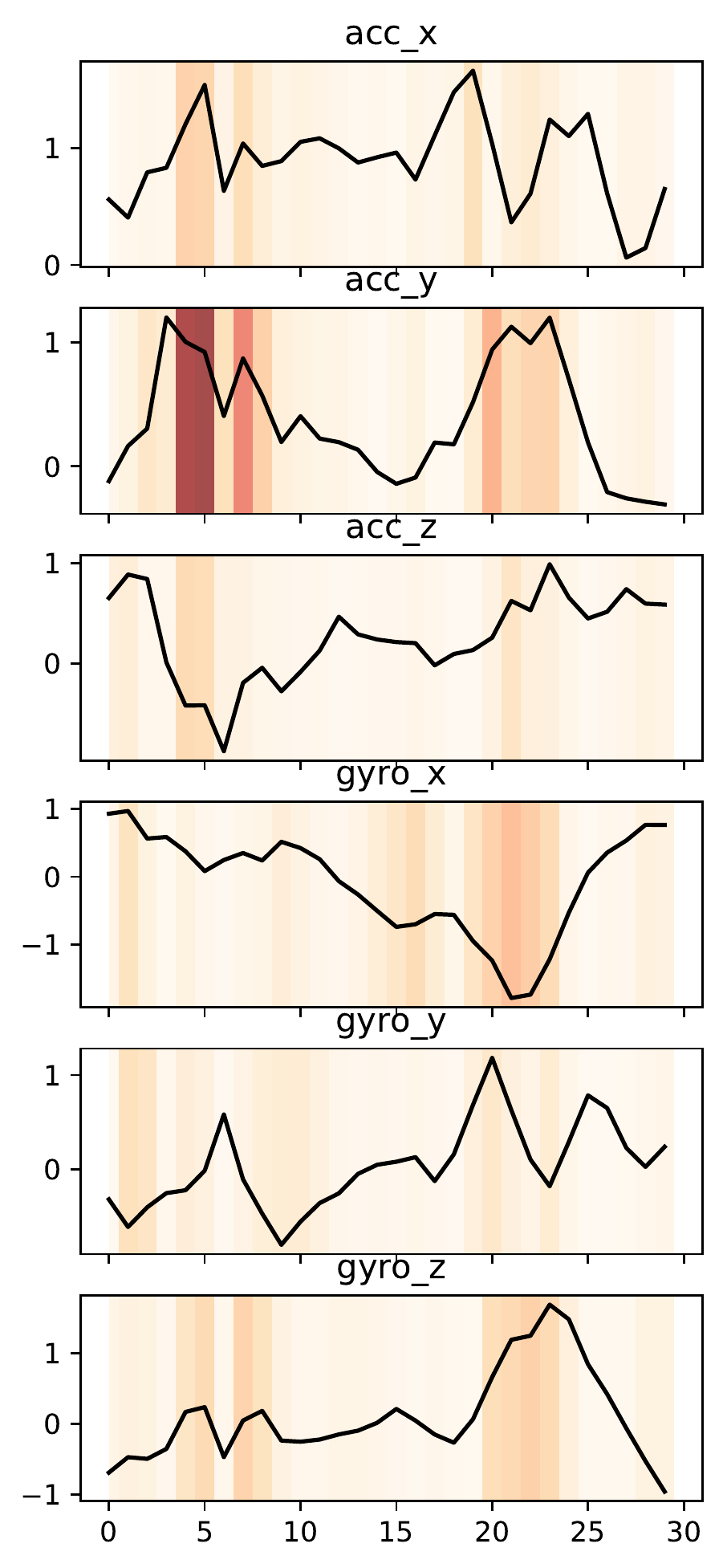}}
%  \vspace{2.0cm}
  \centerline{(c) Sup: Down}\medskip
\end{minipage}

\hfill
\begin{minipage}[b]{.32\linewidth}
  \centering
  \centerline{\includegraphics[width=2.2cm]{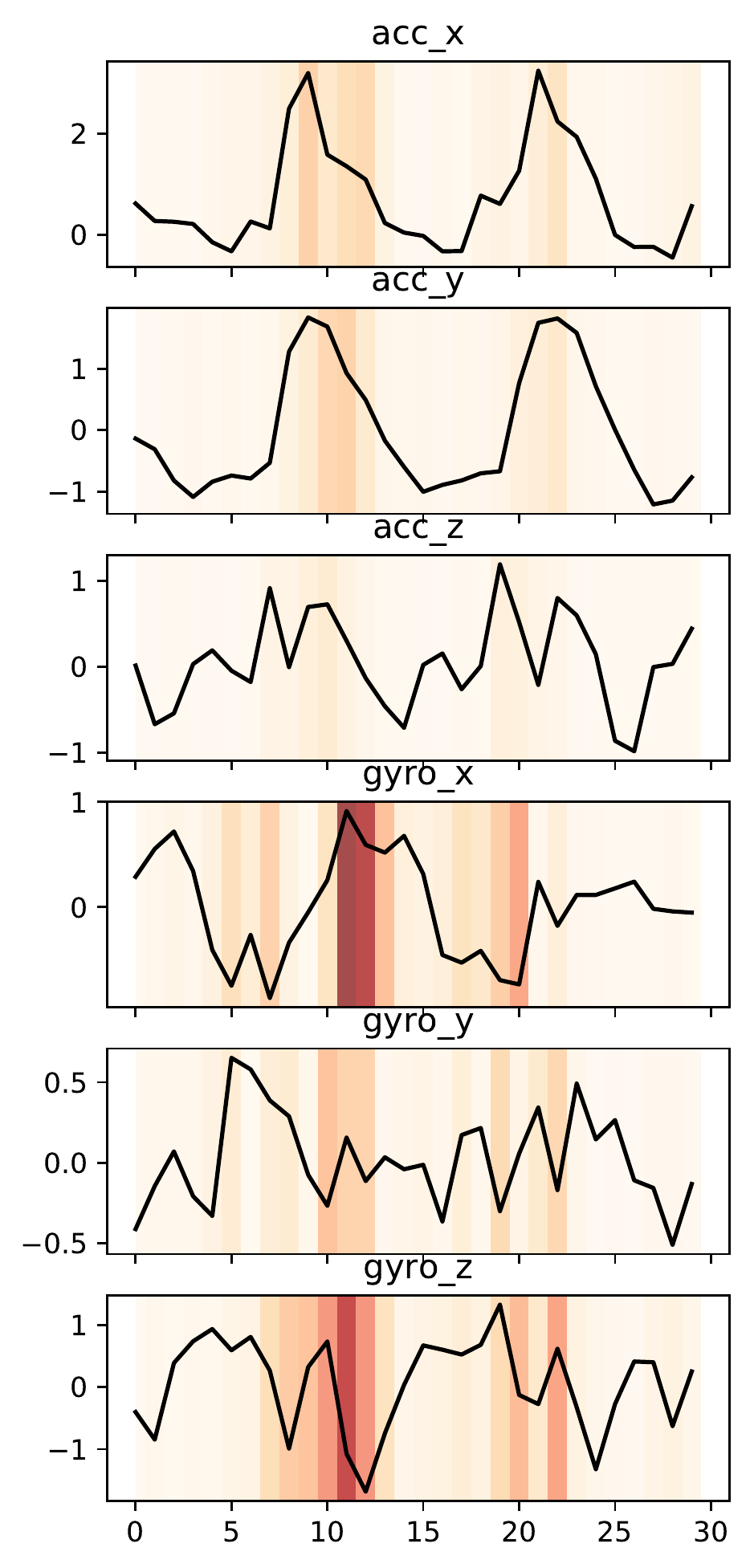}}
%  \vspace{1.5cm}
  \centerline{(d) SimCLR: Jump}\medskip
\end{minipage}
\hfill
\begin{minipage}[b]{.32\linewidth}
  \centering
  \centerline{\includegraphics[width=2.2cm]{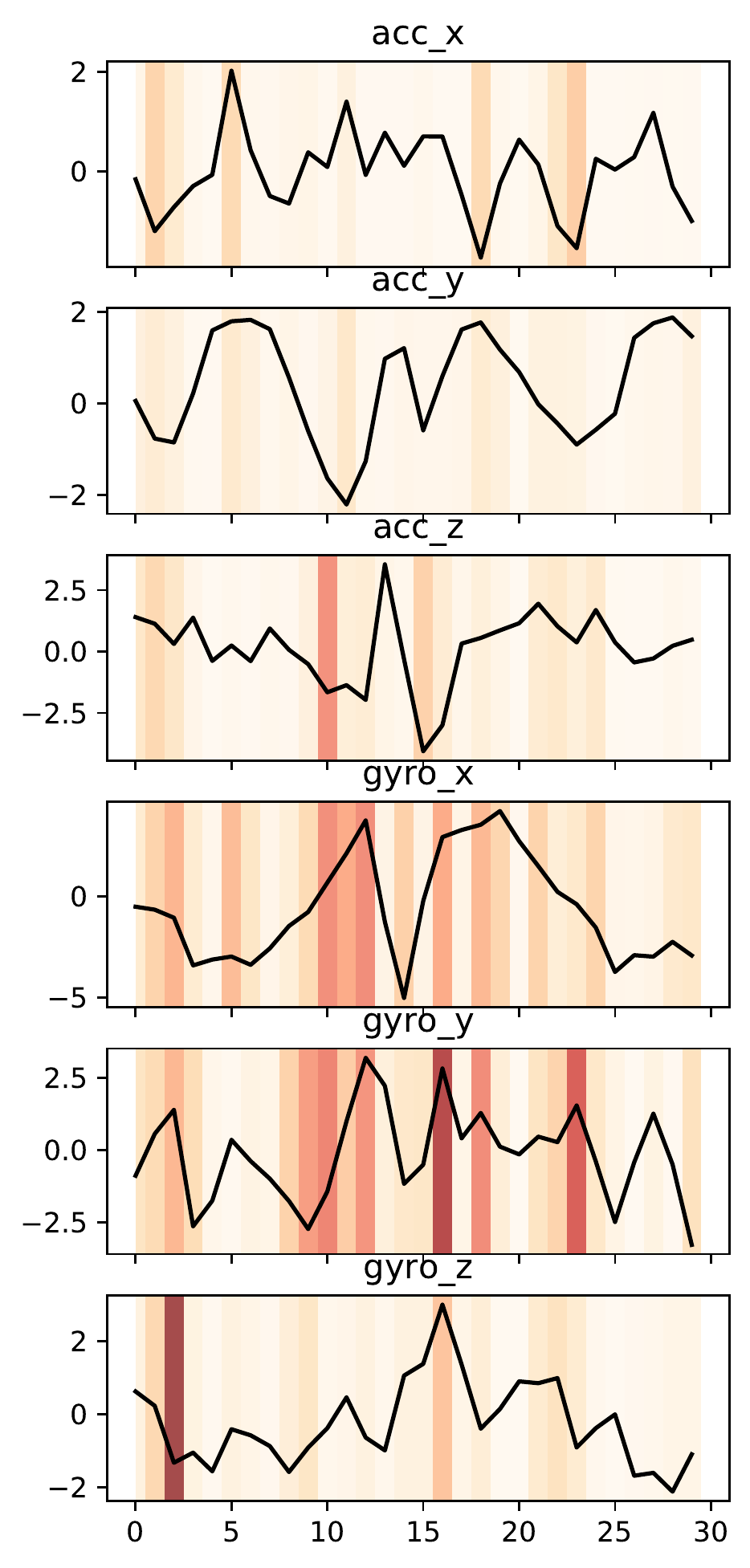}}
%  \vspace{2.0cm}
  \centerline{(e) SimCLR: Jog}\medskip
\end{minipage}
\hfill
\begin{minipage}[b]{.32\linewidth}
  \centering
  \centerline{\includegraphics[width=2.12cm]{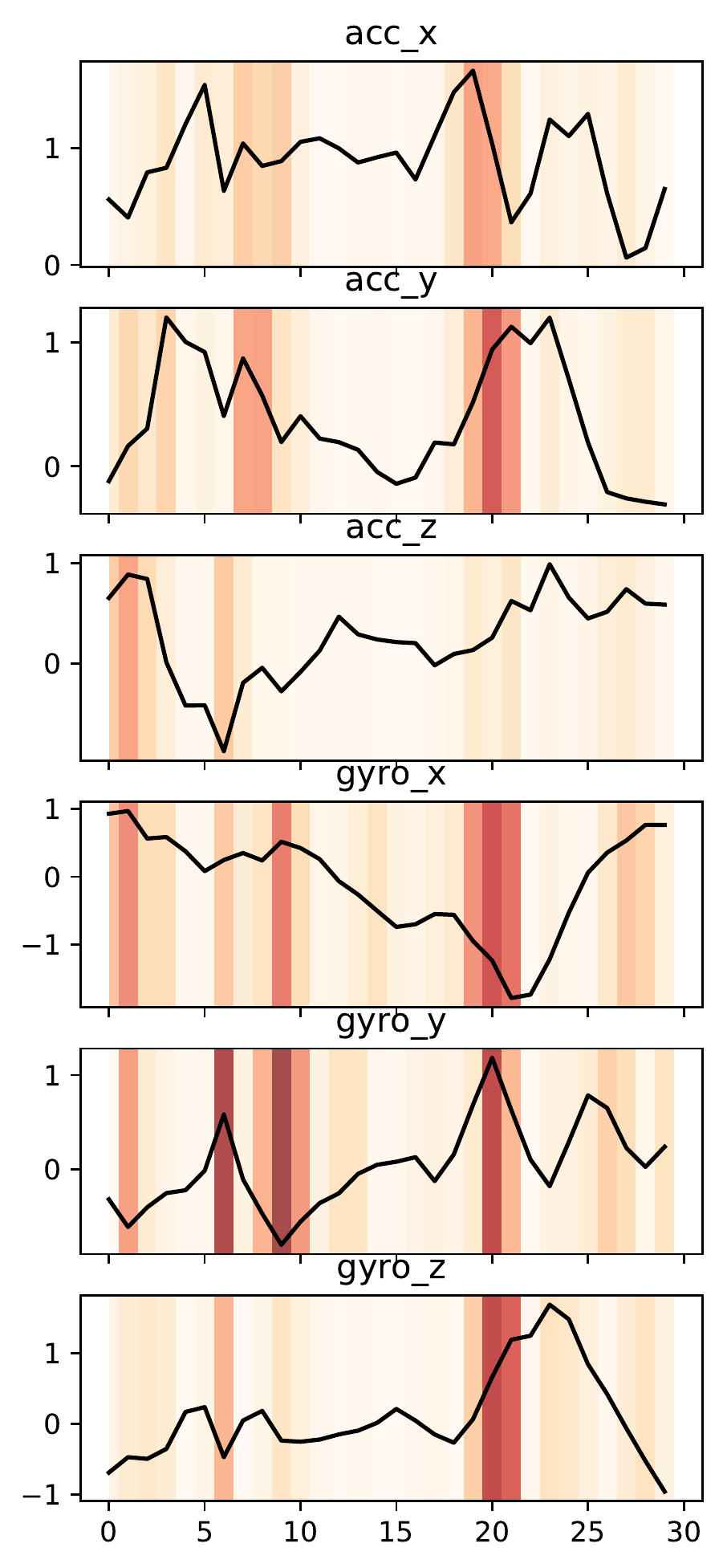}}
%  \vspace{1.5cm}
  \centerline{(f) SimCLR: Down}\medskip
\end{minipage}

\caption{Local attributions generated for jumping activity from MobiAct using Guided Grad-CAM. The intensity of the red color is proportional to the attribution value. Jump, Jog, and Down stand for Jumping, Jogging, and Going Downstairs activities, respectively. Acc\_x, acc\_y, acc\_z, and gyro\_x, gyro\_y, gyro\_z refer to three channels of the accelerometer and gyroscope, respectively.}
\label{fig:local_guided_gradcam}
\end{figure}

We exploit Guided Grad-CAM \cite{selvaraju_2017_gradcam} to generate local attributions for each input channel of the three-axial accelerometer and gyroscope (both have x, y, and z-axis). An example of the attributions for three different activities from the MobiAct dataset is shown in Figure \ref{fig:local_guided_gradcam}. As seen, the supervised model mostly relies on accelerometer data, while SimCLR pays more attention to gyroscope channels. 

\begin{figure}[t]
\begin{minipage}[b]{.45\linewidth}
  \centering
  \centerline{\includegraphics[width=4cm]{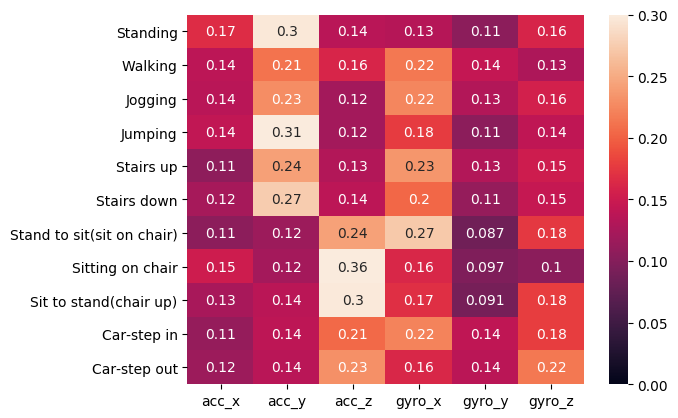}}
%  \vspace{2.0cm}
  \centerline{(a) MobiAct Supervised}\medskip
\end{minipage}
\begin{minipage}[b]{.45\linewidth}
  \centering
  \centerline{\includegraphics[width=4cm]{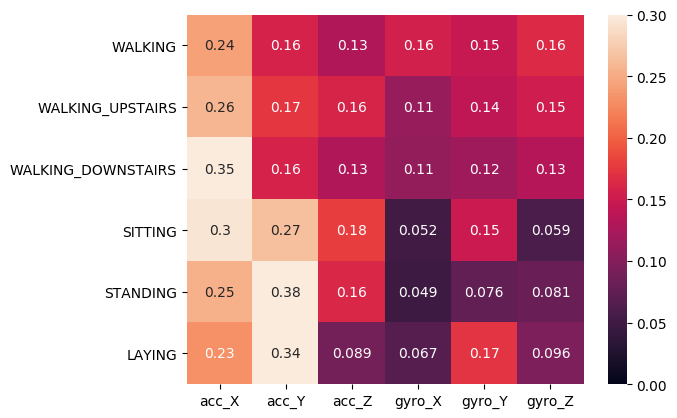}}
%  \vspace{1.5cm}
  \centerline{(b) UCI-HAR Supervised}\medskip
\end{minipage}
\hfill

\begin{minipage}[b]{.45\linewidth}
  \centering
  \centerline{\includegraphics[width=4cm]{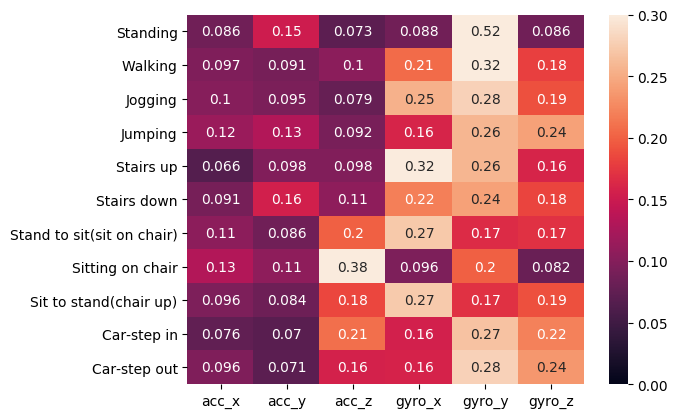}}
%  \vspace{2.0cm}
  \centerline{(c) MobiAct SimCLR}\medskip
\end{minipage}
\begin{minipage}[b]{.45\linewidth}
  \centering
  \centerline{\includegraphics[width=4cm]{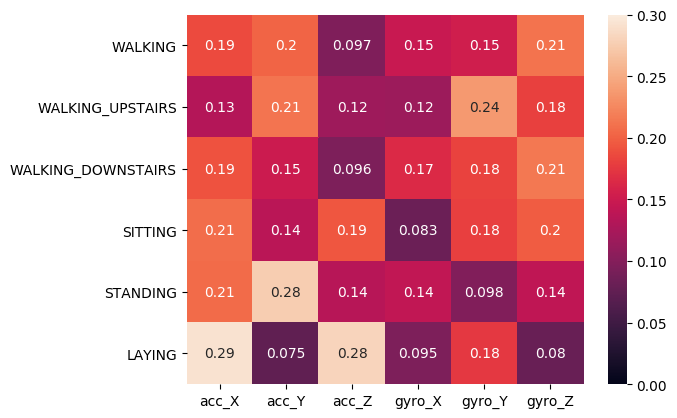}}
%  \vspace{1.5cm}
  \centerline{(d) UCI-HAR SimCLR}\medskip
\end{minipage}
\hfill

\begin{minipage}[b]{.45\linewidth}
  \centering
  \centerline{\includegraphics[width=4cm]{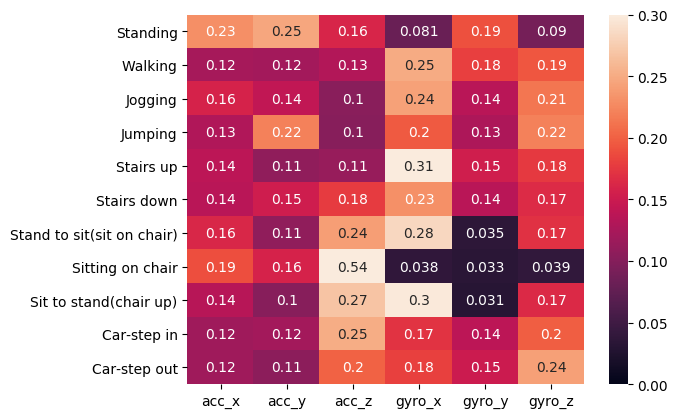}}
%  \vspace{2.0cm}
  \centerline{(e) MobiAct VICReg}\medskip
\end{minipage}
\begin{minipage}[b]{.45\linewidth}
  \centering
  \centerline{\includegraphics[width=4cm]{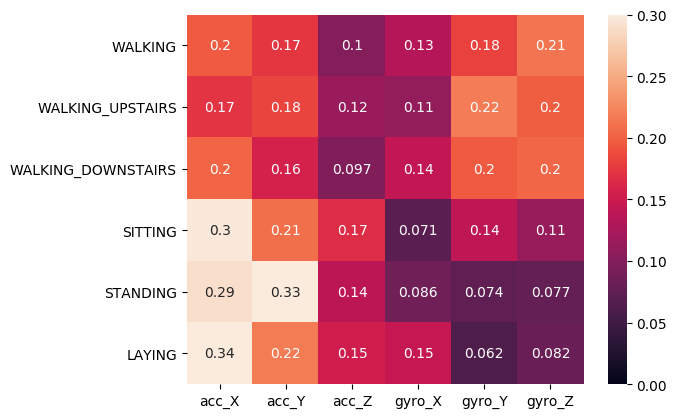}}
%  \vspace{1.5cm}
  \centerline{(f) UCI-HAR VICReg}\medskip
\end{minipage}
\hfill
\caption{Global attribution heatmaps.}
\label{fig:global_gradcam}
\end{figure}

Although local attributions are useful to interpret model decisions given a certain input, they do not indicate global channel importance for each activity. Thus, for each correctly classified input instance in test data, we computed sums of absolute values of attributions per device channel across all time steps and $l_1$-normalized them. As a result, for each input sequence, we have obtained categorical distribution of attributions per device channel. Finally, we average attributions per channel and activity and show them as heatmaps in Figure \ref{fig:global_gradcam}. For both datasets, attribution values calculated for the supervised model are typically higher for the accelerometer channels. In contrast, both SSL methods have similar patterns and can rely on different accelerometer and gyroscope channels depending on activity. What is common for all three frameworks is that they demonstrate the sharper distribution of attributions for most stable and transition activities compared to periodic ones (activity categorization from Table \ref{tab:activity-types}). In order to quantitatively measure the dispersion of attributions obtained for supervised and SSL models, we compute entropy for each distribution and average for all time windows from each activity (shown in Tables \ref{tab:mobiact-entropy} and \ref{tab:ucihar-entropy}). As can be seen from the tables, average entropy values are normally higher for periodic activities meaning their distribution of attributions across input channels is softer. 

\subsection{Representation Probing}
In this work, the probing model is a linear layer with the softmax activation function that is applied on top of the representations pre-trained using one of the SSL or supervised frameworks. The probing model is trained for 10 epochs with a learning rate of $10^{-4}$. The results for subject heterogeneity and activity type probing tasks are shown in Table \ref{tab:probing}. 

\begin{table}[!t]
\centering
\scalebox{0.75}{
\begin{tabular}{rccc}
Activity         & Supervised & SimCLR & VICReg \\ \hline
Standing         &  {2.41}       & \underline{2.01}   & 2.25   \\
Walking          &  \textbf{2.51}       & {2.38}   & 2.45   \\
Jogging          &  {2.5}        & {2.38}   & \textbf{2.47}   \\
Jumping          &  {2.43}       & 2.42   & {2.41}    \\
Stairs up        &  {2.45}       & {2.32}   & 2.4   \\
Stairs down      &  {2.47}       & \textbf{2.46}   & {2.45}   \\
Stand to sit     &  2.37       & {2.4}    & {2.22}   \\
Sitting on chair &  \underline{2.3}        & 2.14   & \underline{1.66}   \\
Sit to stand     &  {2.39}       & {2.39}   & {2.18}   \\
Car-step in      &  {2.48}       & {2.36}   & 2.38   \\
Car-step out     &  {2.44}       & 2.37   & {2.36}   \\ \hline
Overall          &  {2.45}       & {2.24}   & 2.34  
\end{tabular}}
\caption{Values of entropy computed for accumulated attributions per activity on MobiAct. The highest and lowest values per framework are highlighted in bold and underlined, respectively.}
\label{tab:mobiact-entropy}
\end{table}

\begin{table}[t]
\centering
\scalebox{0.75}{
\begin{tabular}{rccc}
Activity           & Supervised & SimCLR & VICReg \\ \hline
Walking            & \textbf{2.53}       & \textbf{2.42}   & \textbf{2.49}   \\
Walking Upstairs   & {2.47}       & {2.4}    & {2.47} \\
Walking Downstairs & {2.39}       & 2.41   & \textbf{2.49}  \\
Sitting            & {2.25}       & {2.32}   & 2.3  \\
Standing           & \underline{2.14}       & {2.39}   & \underline{2.17}  \\
Laying             & {2.28}       & \underline{2.17}   & 2.2  \\ \hline
Overall            & {2.34}       & {2.38}   & 2.35 
\end{tabular}}
\caption{Values of entropy computed for accumulated attributions per activity on UCI-HAR. The highest and lowest values per framework are highlighted in bold and underlined, respectively.}
\label{tab:ucihar-entropy}
\end{table}

\noindent{\textbf{Subject heterogeneity.}} As described in Section \ref{sec:meth_xai}, the subject heterogeneity task evaluates the subject invariance of the learnt representations. Specifically, we merge all the time windows from the train, validation, and test subjects and randomly divide them into 5 folds for cross-validation. The probing model in this case is trained to predict subject labels. As can be seen from Table \ref{tab:probing} the subject heterogeneity score is significantly higher for SSL models. The low subject heterogeneity score means that a model generalizes well and representations of the same activity from different subjects are close and homogeneous. In contrast, SSL can be a good choice when subjects have to be re-identified or features should allow personalization, yet satisfactory activity recognition performance should be maintained.

\noindent{\textbf{Activity type.}} The activity type probing task aims to evaluate if the pre-trained representations contain properties representing the nature of an activity. The activities in the datasets have been categorized as periodic, stable, and sporadic as shown in Table \ref{tab:activity-types}. The evaluation scenario employed in this task is leave-one-activity-out, i.e. we hold out all time windows corresponding to one activity from each category. These data are later used as a test set, while the remaining activities are employed as a training set. The probing model in this case is trained to predict activity type labels (periodic, stable, or sporadic) and evaluated on unseen activities. The experiment is conducted for all possible combinations of activities in the test set (40 and 9 runs for MobiAct and UCI-HAR, respectively). As can be seen from Table \ref{tab:probing}, the supervised model better encodes information about the nature of activities achieving performance metrics that are on average higher for both datasets, namely 50.41 and 87.6\% F1-score for MobiAct and UCI-HAR.
\begin{table}[!t]
\centering
\scalebox{0.75}{
\begin{tabular}{cr|cc}
{Dataset}                   & {Approach} & {SH (F1-score)} & {AT (F1-score)} \\ \hline
                          & {Sup}      & {59 $\pm$ 0.28}                      & {\textbf{50.41 $\pm$ 5.26}}  \\
                         & {SimCLR}   & {\textbf{75.87 $\pm$ 0.45}}          & {39.99 $\pm$ 3.48}           \\
\multirow{-3}{*}{{MobiAct}} & VicReg                          & 67.31 $\pm$ 0.33                                          & 38.59 $\pm$ 2.84                                  \\ \hline
                          & {Sup}      & {6.87 $\pm$ 0.26}                    & {\textbf{87.6 $\pm$ 21.1}}   \\
                         & {SimCLR}   & {48.7 $\pm$ 0.77}           & {75.56 $\pm$ 10.54}          \\
\multirow{-3}{*}{{UCI-HAR}} & VicReg                          & \textbf{57.81 $\pm$ 0.62}                                          & 55.38 $\pm$ 7.67                                 
\end{tabular}}
\caption{Representation probing results: average values of F1-scores with margins of error for 95\% confidence intervals. SH and AT stand for subject heterogeneity and activity type.}
\label{tab:probing}
\end{table}

\subsection{Limitations}
\label{sec:limitations}
There are limitations that we would like to highlight for future work on interpreting supervised and self-supervised sensor-based HAR models. \textbf{(1)} There is no clear link between observations found in the current combination of occlusion and saliency-based methods. In the occlusion experiments, the largest drop in performance is observed for stable and transition activities. Meanwhile, we also highlighted that activities belonging to these types do normally have sharper distributions of saliency attributions (according to the entropy values and heatmaps). However, there is no clear evidence that these observations are connected. \textbf{(2)} There is an ongoing discussion in the literature \cite{adebayo_2018_sanity, yona_2021_RevisitingSC} on whether the Guided Backpropagation method (which is a part of Guided Grad-CAM used in this study) generates saliency maps that are not sensitive to the underlying model. In turn, the plain Grad-CAM creates averaged attribution maps across all input channels, and hence, it cannot be used to compare the contributions of each channel. Thus, future works could also compare Guided Grad-CAM to other saliency methods, such as DeepLIFT \cite{shrikumar2017learning} or Integrated Gradients \cite{sundararajan2017axiomatic}. \textbf{(3)} Open-source datasets widely used in the sensor-based HAR literature, including UCI-HAR and MobiAct, contain a relatively small number of basic activities. Besides, the activity type probing task in this study uses three generic activity groups. A natural progression of this work is to probe representations on a dataset with the large number of diverse activities separated into more groups to assess how supervised and SSL models encode similar types of activities.
\section{Conclusion}
In this paper, we implement three families of explainability approaches for sensor-based HAR to evaluate representations learnt in supervised and self-supervised settings. Based on the occlusion experiments, the SSL frameworks might be a better choice in a realistic scenario when training data is cleansed but test data is corrupted. Moreover, using Guided Grad-CAM, we show what channel supervised and SSL approaches typically pay attention to recognize various activities. Finally, according to the probing experiments, the SSL representations are more suitable for applications that enable personalization, while the supervised models produce more homogeneous representations per activity class. Besides that, the supervised models better encode properties related to the nature of input activities.

Future work on the explainability of SSL frameworks for HAR might focus on addressing the limitations highlighted in Section \ref{sec:limitations}. Besides, it would be interesting to conduct a similar analysis on datasets with multiple devices and multimodal datasets in order to analyze supervised and SSL representations on a per body part level.

\section*{Disclaimer}
Dr. Stylianos Asteriadis is currently working at the European Commission. The information and views set out in this article are those of the authors and do not necessarily reflect the official opinion of the Institution.

{\small
\bibliographystyle{ieee}
\bibliography{egbib}
}

\end{document}